# Imperialist Competitive Algorithm with Independence and Constrained Assimilation for Solving 0-1 Multidimensional Knapsack Problem

Ivars Dzalbs, Tatiana Kalganova[1], Ian Dear

1 Brunel University London, Kingston Lane, Uxbridge, UB8 2PX, UK
Tatiana.Kalganova@brunel.ac.uk

**Abstract.** The multidimensional knapsack problem is a well-known constrained optimization problem with many real-world engineering applications. In order to solve this NP-hard problem, a new modified Imperialist Competitive Algorithm with Constrained Assimilation (ICAwICA) is presented. The proposed algorithm introduces the concept of colony independence – a free will to choose between classical ICA assimilation to empire's imperialist or any other imperialist in the population. Furthermore, a constrained assimilation process has been implemented that combines classical ICA assimilation and revolution operators, while maintaining population diversity. This work investigates the performance of the proposed algorithm across 101 Multidimensional Knapsack Problem (MKP) benchmark instances. Experimental results show that the algorithm is able to obtain an optimal solution in all small instances and presents very competitive results for large MKP instances.

## 1. Introduction

The Multidimensional Knapsack Problem (MKP) is a well-known constrained optimization problem, that has multiple real-world engineering applications, such as cutting stock [1], distributed computing resource allocation [2], cargo loading [3], satellite management [4], project selection [5] and capital budgeting [6]. The MKP is an extension of the 0-1 knapsack problem, where items have weight vectors in multiple dimensions. The goal is to maximize the total profit by putting items into knapsacks while satisfying weight capacity constraints across all dimensions. MKP is formulated in (1) [7].

$$\begin{aligned} \max \quad & \sum_{j=1}^{n} p_j s_j \\ \text{subject to:} \quad & \sum_{j=1}^{n} w_{ij} s_j \leq W_i & \forall i \in \{1, \dots, m\} \\ & x_j \in \{0,1\} & \forall j \in \{1, \dots, n\} \end{aligned} \quad (1)$$

where every item $j$ in the list of $n$ items ($j = 1 \dots n$) has a profit $p_j$ and weight $w_{ij}$ associated to an $m$-dimensional weight vector ($i = 1 \dots m$), that tries to satisfy a capacity constraint $W_i$ in that dimension. Variable $s_j$ indicates whether the item is selected and included in solution. Capacities, weights and profits are assumed to be positive. Being an NP-hard problem with practical applications, many different approaches have been proposed for solving MKP, which can be divided into two groups – exact, deterministic, single-solution based algorithms and stochastic population/meta-heuristic based algorithms, with this paper focusing on the latter approach.

Comprehensive literature review of solving MKPs was provided by [8] and a more recent MKP overview by [9] summarizes algorithms used for solving MKP. This paper focuses on the state-of-the-art population and meta-heuristic algorithms used for solving MKP instances, such as ant colony optimization ([10], [11]), various types of genetic algorithms ([12],[13],[14]), evolutionary algorithms ([15],[16], [17]), variations of particle swarm optimization algorithm ([18],[19]), binary harmony search [20], binary cuckoo search algorithm[21], whale optimization algorithm [22] and etc. Most of the research in population-based algorithms focuses on small MKP instances with $n \leq 100$, while only few explore large instances with $n = 500$ and above. This paper tries to cover both small and large instances of MKP for a complete study using an Imperialist Competitive Algorithm (ICA).

The main contributions of this work are as follows:
- Investigation of the Imperialist Competitive Algorithm (ICA) for the first time applied to large MKP.
- Proposal of novel generic ICA implementation for solving constrained optimization problems. The proposed approach called ICA with Independence and Constrained Assimilation (ICAwICA) combines classical ICA assimilation and mutation operators by modified assimilation process with repair mechanism.
- Computation results and comparisons of 101 commonly used MKP benchmarks, of which 30 are very small instances, with $n < 100$, 60 are small instances, with $n = 100$ and 11 large benchmark instances with $n$ up to 2500.

## 2. The Imperialist Competitive Algorithm

The Imperialist Competitive Algorithm (ICA) was first introduced in 2007 [23] for solving continuous optimization problems and since been a growing field of interest for many researchers in various engineering disciplines – scheduling, assembly line balancing, facility layout optimization, computer engineering and other areas of industrial engineering [24].

Like many other population algorithms, ICA starts its search by generating a random initial population where each individual of the population represents a country. Countries within ICA can be thought of as chromosomes in a genetic algorithm. The initial population is separated into multiple groups (so called empires). Strongest countries become imperialist within the empire and weakest - their colonies. Each colony within empire moves closer to their imperialist in the form of assimilation operator. In order to provide diversity amongst countries, a revolution operator (mutation in GA) is implemented. If at any point a colony becomes stronger than its imperialist then the two countries are swapped, such that imperialist is the strongest country in the empire. The search follows an iterative process, where after each iteration the weakest colony within the weakest empire is assigned to one of the stronger empires – following the imperialist competition process. An empire is eliminated once it contains no more colonies. The search usually continues until the termination criteria are met. Ideally, the search is terminated once all empires are eliminated and only one, the best, empire remaining.

Multiple variations of ICA have been previously proposed, for instance, both [25] and [26] implemented ICA with attraction and repulsion mechanism to assist the search with great success. Moreover, [27] explored variations of imperialist competition, where the strongest imperialist gets excluded and also proposed diversification operator to escape local optima. Furthermore, to improve search convergence, many ICA hybrids with other algorithms have been explored. ICA hybrids, such as ICA with teaching-learning based optimization [27] for solving multi-stage supply chains, combination of ICA and simulated annealing algorithm for solving flexible job scheduling problem [28], as well as ICA with particle swarm optimization for power resource management [29].

## 3. The Imperialist Competitive Algorithm with Independence and Constrained Assimilation

The classical ICA does not implement any form of local search and therefore may get stuck in local optima before converging to global best solution [30]. Different approaches for solving this problem have been proposed in literature, such as simulated annealing-like processes in [31] for solving flexible job-shop problems, as well as applying local search for sub-populations of ICA [32]. Work in this paper proposes a modified ICA, where the local search process is performed in terms of both an Independence operator and a Constrained Assimilation (ICAwICA). The flowchart for both classical ICA and ICAwICA is shown in Fig. *1*, with red indicating the changes.

The proposed ICAwICA follows the classic ICA [23] principles for both empire initialization and empire competition, however, assimilation and revolution operators are replaced with a constrained assimilation and repair mechanism. Furthermore, in the classical ICA each of the colonies within an empire are moving closer to the imperialist within that empire, while in ICAwICA all colonies are given a free choice to move closer to any of the imperialists of other empires (independence), as long as it improves the country's well-being (associated cost). Therefore, at each iteration, colony $k$ has a probability based on a uniform distribution ($rand$) of either moving closer to their own empire's imperialist or to move closer to any other imperialist $j$, determined by $independanceRate$ (0-1.0). Moreover, this process is repeated $N_{localIter}$ times for each colony to explore more search space around its position in the form of local search. Pseudo code of the ICAwICA is shown in Fig. *2*.

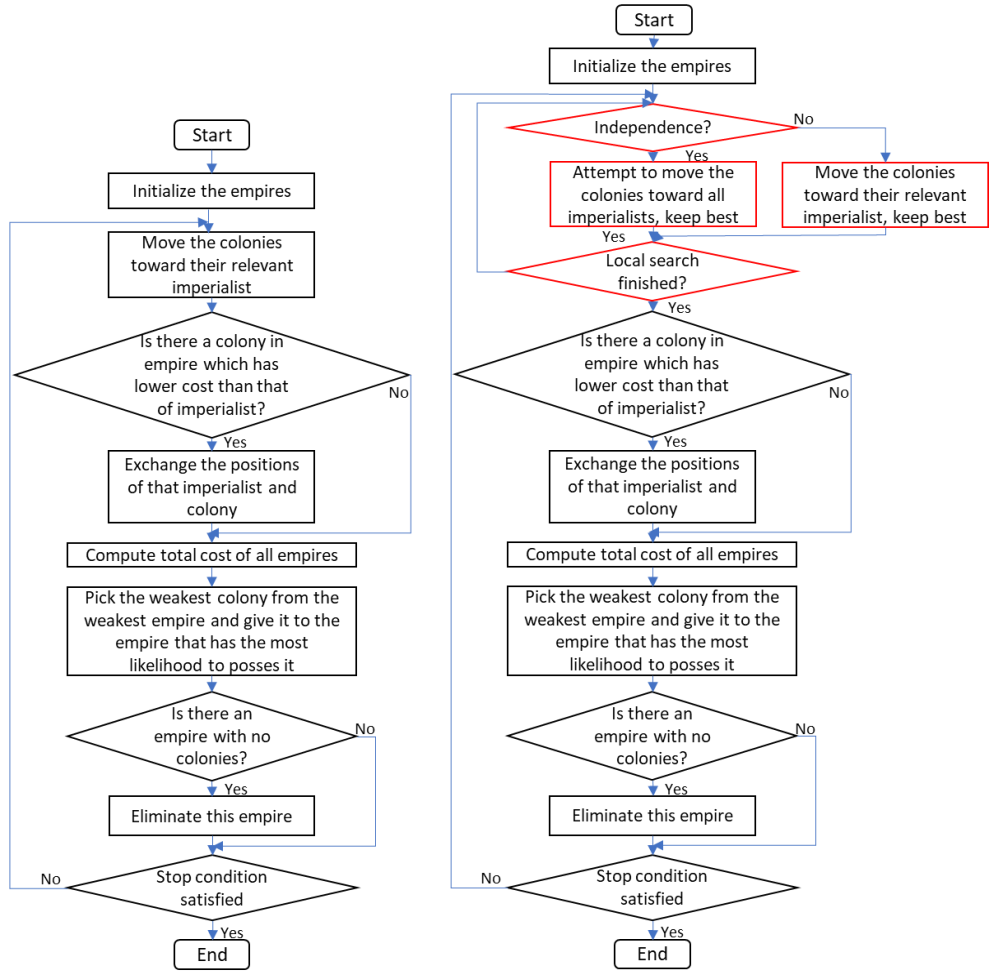

**Fig. 1.** Flowchart of classical ICA [23] (to the left) and the proposed ICAwICA (on the right), with red indicating the changes.

### 3.1. Constrained Assimilation

Classical ICA was first developed for continuous maths problem with simple assimilation processes [23], ICA has since been applied to multiple binary problems, such as feature selection [33][34], content-based-image retrieval (CBIR) [35] and single dimensional 0-1 knapsack problems [36]. However, binary assimilation approaches cannot always be extended to other discrete, non-binary problems. Furthermore, most ICA discrete assimilation implementations follow simple genetic-algorithm-like crossover operations, where the chromosomes are expected to be of equal size [37] [38]. The proposed Constrained Assimilation (CA) process does not require equal chromosome/solution size and is extendable to other constrained discrete problems. CA exploits the fact that two solutions cannot always be merged without violating constraints. Therefore, CA builds a new incomplete solution from the two donor solutions/countries (colony and imperialist) according to the assimilation rate and finishes the solution by a repair mechanism.

There are multiple ways to implement the solution repair mechanism - based on heuristics, existing solution population, sequence based [39] etc. The simplest repair mechanism is - scanning through all possible entries and trying to add them to the solution without violating constraints (used in this paper). Furthermore, this incomplete solution repair enables diversity without an explicit revolution operator like classic ICA. Although more computationally expensive than simple assimilation, this approach has potential for broad applications and generalization, as it does not depend on two solutions having the same size nor problem-specific assimilation or repair mechanism.

A CA example is provided in Fig. *3* where both colony and imperialist are assimilated, with bold integer values corresponding to solution entries (item indices in MKP case) that are passed to the new country, determined by assimilation rate. In this simple example, a 50% assimilation rate of $N_{solutionSize}$ is used to build the new country. Due to constraints, not all solution entries can be added to the new country and hence the

```
1. Initialize ICA parameters.
2. Create the popular randomly.
3. Initialize empires:
   for i = 1 to N_{population}
           Compute the cost function C_i;
           Sort the computed cost C_i in descending order for the entire population;
           Select N_{imperialists} out of N_{population};
           Normalize the cost of each imperialist C_n;
           Calculate the normalized power of each imperialist P_n;
           Assign remaining countries N_{countries} to the imperialists;
   end loop
do
   4. Assimilation and local search process for ICAwICA
     for k = 1 to N_{colonies}
       for l = 1 to N_{localIter}
         if rand < independanceRate
           for j = 1 to N_{imperialists}
               assimilate colony k closer to j
               if cost for new position is less than original position
                   keep assimilated position
               else
                   discard and move back to original position
               endif
           end loop
         else
           assimilate colony k closer to empire's Imperialist
         endif
       end loop
     end loop
     for j = 1 to N_{imperialists}
         if the cost of any colony is less than cost of imperialist
             exchange the position of the colony and imperialist;
         endif
         Pick the weakest colony (colonies) from the weakest empire and assign it to the
         empire with highest probability to possess it;
     end loop
   5. Elimination process;
       If there is imperialist with no colonies
           eliminate the imperialist;
       endif
while stopping condition not met;
```

**Fig. 2.** The pseudo-code for new assimilation and local search method for ICAwICA

solution is in an incomplete state. The repair process iterates over all possible solution entries and fills the gaps while complying with constraints. As can be seen in Fig. *3*, new entries (index 1 and index 3) were introduced to the solution after repair that were not in any of the donor countries.

Let us consider in detail the assimilation process shown in Fig. *3*. The colony solution is shown in blue and the imperialist in yellow, with the newly generated country $nc$.

|  | $N_{solutionSize}$ | | | | | | |
|---|---|---|---|---|---|---|---|
| Colony | **2** | 9 | 8 | **6** | **7** | 13 | 5 |
| Imperialist | 12 | **19** | **5** | 12 | 2 | 6 | 15 | **8** |
| New country $nc$ | 2 | 19 | 5 | 6 | 7 |  |  | 8 |
| $nc$ after repair | 2 | 19 | 5 | 6 | 7 | 1 | 3 | 8 |

**Fig. 3.** Imperialist and colony constrained assimilation process with solution repair. Entries to be passed onto new country represented in bold. With integer values corresponding to solution entries (item indices in MKP case).

## 4. Experimental result

In this section the proposed algorithm performance is evaluated by extensive computational experiments on classical MKP benchmark instances and compared to the current state-of-the-art algorithms.

### 4.1. Benchmark instances

Multidimensional knapsack problem instances were chosen because of their availability, ease of implementation and the common use as benchmarks across the research community. ICAwICA was tested across 101 popular benchmark instances, all available from the compiled library in [40].

The simplest benchmarks used in this paper are derived from the WEISH dataset, containing 30 problems with the number of items ranging from 30 to 90 and with 5 knapsacks each. Furthermore, another common MKP benchmark - OR-library, generated by Chu and Basley in [41], was selected. Each group of OR set instances (9 in total) contains 30 problems with 3 different tightness ratios, set to 0.25 for first 10 instances, 0.5 for next 10 instances and 0.75 for last 10 instances. 60 problems were selected from the set - 30 with 100 items, 5 knapsacks and 30 with 100 items, 10 knapsacks. In order to explore the performance of the proposed algorithm across range of datasets, large MKP instances, generated by Glover and Kochenberger (GK) [42], were also selected. The GK dataset contains 11 instances with number of items ranging from 100 to 2500 with 15 to 100 knapsacks each and provides wide spectrum of complexity.

### 4.2. Experimental setup

The proposed ICAwICA algorithm was implemented in C++ using the Visual Studio 2019 (v142) compiler. The computation was performed on a workstation with AMD Threadripper 2990WX processor (3.0 GHz, 64GB ram), running Windows 10 Pro operation system.

Similar to classic ICA, ICAwICA also has multiple algorithmic hyper-parameters that were empirically set and are as follows: total number of countries $N_{population}$ is set to 4096 for all instances with number of items $n < 500$ and 512 for all instances with $n \geq 500$. Out of all countries, 40% are initialized as imperialists $N_{imperialists}$. Local iterations $N_{localIter}$ is set to 3. Assimilation rate β set to 0.5; coefficient associated with average power of empire's colonies $\xi$ set to 0.05; $independanceRate$ set to 0.7 (70% probability of independence).

Due to constrained computing resources, limited time and a large set of problem set, termination criteria of stagnation were implemented, where the search terminates if no improvement has been made to the best solution for ε number of iterations. For problem instances with $n < 500$, ε is set to $0.1n$, and for MKP instances with h $n \geq 500$, $\varepsilon = n$. Due to the stochastic nature of the algorithm, 30 independent runs were computed for each problem instance. Best and average profit as well as average time in seconds $t_{avg}(s)$ required to reach such profit were recorded for all 101 problem instances.

### 4.3. Sensitivity analysis of $independanceRate$

The newly implemented mechanism of colony independence was tested by altering the $independanceRate$ parameter from 0 to 1, with 0.1 increments. OR10.100 problem set was chosen for the comparison. The average error percentage across all 30 OR10.100 instances was calculated in (2) and the experimental results are summarized in Table 1.

$$Average\ error\ (\%) = \frac{1}{n}\sum_{i=1}^{n}\frac{o_i - p_i}{o_i} * 100\% \qquad (2)$$

where $o_i$ is the optimal profit for the instance $i$, and $p_i$ – achieved best or average profit on said instance.

**Table 1.** Sensitivity analysis of Independence rate as an average error % for OR10.100 instances across 30 independent runs. With 0 representing ICA with no independence operator. $t_{avg}(s)$ representing the average time to converge to best solution.

|  | Independence rate | | | | | | | | | | |
| --- | --- | --- | --- | --- | --- | --- | --- | --- | --- | --- | --- |
|  | 0 (no) | 0.1 | 0.2 | 0.3 | 0.4 | 0.5 | 0.6 | 0.7 | 0.8 | 0.9 | 1 |
| Average error % | 5.1944 | 0.0283 | 0.0153 | 0.0081 | 0.0061 | 0.0041 | 0.0039 | **0.0026** | 0.0027 | 0.0029 | 0.0031 |
| $t_{avg}(s)$ | 0.14 | 2.13 | 2.85 | 3.64 | 4.27 | 5.10 | 5.68 | 6.51 | 7.21 | 7.98 | 8.74 |

Table 1 shows a clear improvement in the introduction of the Independence operator within ICA. Compared to ICA with no independence (independence rate of 0) and ICA with independence rate greater than 0, the average error across all OR10.100 instances reduced by factor of 183 (5.1944% and 0.0283% respectively). However, there is also a time penalty associated with doing the extra work of assimilating to all imperialists compared to a single imperialist, with average time to reach final solution increasing from 0.14 seconds to 2-8 seconds. Best average error was achieved with the Independence rate at 0.7 and hence been adopted for use throughout all further comparisons.

**4.4. Comparisons to the state-of-the-art.**

To evaluate performance of the proposed algorithm, 12 state-of-the art population-based/ heuristic algorithms were compared across 101 common MKP instances. Benchmark instances are described in Section 4.1 and experimental setup algorithmic hyperparameters with termination criteria in Section 4.2.

First, comparison was performed on simple WEISH instances, where most algorithms in the literature can achieve optimum solution, therefore performance is measured in terms of the success rate (how many times algorithm was able to achieve optimum) or in terms of the average error percentage error (2) across all instances. For the comparison, the six best performing algorithms were selected from the literature, which include Ant Colony Optimization with Dynamic impact (ACOwD) described in [11], Improved Whale Optimization Algorithm (IWOA) [22], two variations of binary differential search TE-BDS and TR-BDS proposed in [43], and two implementations of Particle Swarm Optimization (PSO) with self-adaptive check and repair - SACRO-CBPSOTVAC and SACRO-BPSOTVAC [19].

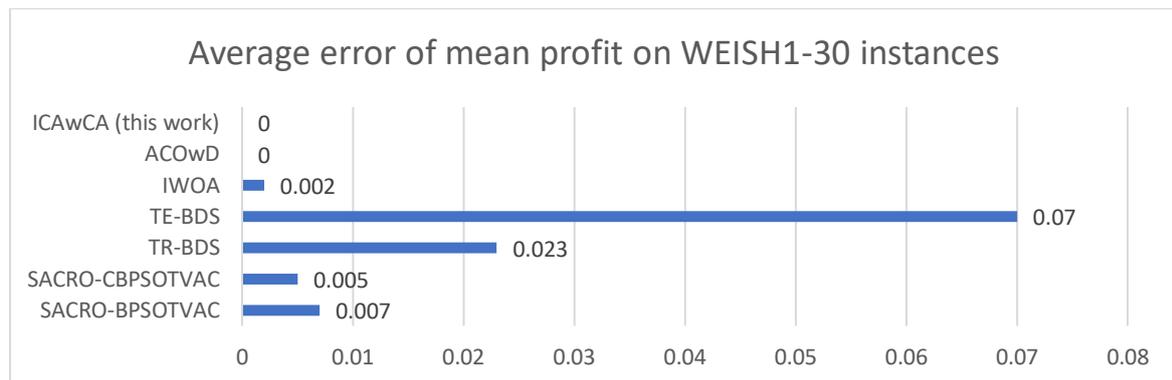

**Fig. 4.** The average error of the mean profit across all WEISH (1-30) instances. Average of 30 independent runs.

Results in Fig. 4 show that all compared algorithms can reach the optimal solution in most cases, however, only 2 of them ICAwICA and ACOwD are able to do it consistently across 30 runs. ICAwICA achieved the optimal solution at the first iteration every time and on average took 1.5 seconds.

Next, 60 OR-library instances were compared with state-of-the-art Two-phase tabu-evolutionary algorithm (TPTEA) [15], Quantum Particle Swarm Optimization technique with local search and repair mechanism called QPSO* [18] and SACRO-CBPSOTVAC, SACRO-BPSOTVAC [19]. Results in both Table 2 and Table 3 are structured as follows. The first two columns show instance name and the absolute value of the optimal solution. The absolute error is derived by subtracting the score achieved from the optimal value. Comparisons are made for both average absolute error and best absolute error achieved within the 30 runs. The average time to achieve solution $t_{avg}(s)$ is added for reference. With bold representing the best error of zero.

ICAwICA was able to achieve the optimal solution for all 60 instances. In terms of best absolute error, ICAwICA consistently outperformed both SACRO-CBPSOTVAC and SACRO-BPSOTVAC and performed just as well, when compared to TPTEA and QPSO*. In terms of average absolute error, ICAwICA consistently achieved an optimal solution in 20 out of 30 OR5.100 instances and 21 out of 30 OR10.100 instances, outperforming SACRO-CBPSOTVAC and SACRO-BPSOTVAC, but falling behind both TPTEA and QPSO*. It is worth noting that the average time to solution $t_{avg}(s)$ almost doubles when solving instances with tightness ratio of 0.5 and 0.75 compared to 0.25, this is due to the extra constraint checks CA has to perform when creating a new solution.

**Table 2.** Algorithm comparison results for small OR-library instances with n= 100 and m = 5. With bold corresponding to the best absolute error. Average derived from 30 independent runs.

| Problem | | Average absolute error | | | | Best absolute error | | | | ICAwICA (this work) | | |
|---|---|---|---|---|---|---|---|---|---|---|---|---|
| Instance | Optimum | TPTEA [15] | SACRO-BPPSO-TVAC [19] | SACRO-CBPPSO-TVAC [19] | QPSO* [18] | TPTEA [15] | SACRO-BPPSO-TVAC [19] | SACRO-CBPPSO-TVAC [19] | QPSO* [18] | Avg | Best | $t_{avg}(s)$ |
| 5.100.0 | **24381** | **0.0** | 85.0 | 79.3 | **0.0** | **0** | 38 | 38 | **0** | **0.0** | **0** | 4.1 |
| 5.100.1 | **24274** | **0.0** | 120.5 | 120.7 | **0.0** | **0** | **0** | **0** | **0** | **0.0** | **0** | 4.1 |
| 5.100.2 | **23551** | **0.0** | 40.8 | 41.3 | **0.0** | **0** | 13 | 13 | **0** | **0.0** | **0** | 4.3 |
| 5.100.3 | **23534** | **0.0** | 101.1 | 76.4 | **0.0** | **0** | 7 | 7 | **0** | **0.0** | **0** | 5.2 |
| 5.100.4 | **23991** | **0.0** | 58.5 | 66.7 | **0.0** | **0** | **0** | 25 | **0** | **0.0** | **0** | 4.5 |
| 5.100.5 | **24613** | **0.0** | 96.6 | 66.8 | **0.0** | **0** | 12 | 12 | **0** | **0.0** | **0** | 3.9 |
| 5.100.6 | **25591** | **0.0** | 150.9 | 140.0 | **0.0** | **0** | **0** | **0** | **0** | **0.0** | **0** | 3.9 |
| 5.100.7 | **23410** | **0.0** | 88.2 | 61.0 | **0.0** | **0** | **0** | **0** | **0** | **0.0** | **0** | 4.2 |
| 5.100.8 | **24216** | **0.0** | 52.5 | 67.5 | **0.0** | **0** | 12 | **0** | **0** | **0.0** | **0** | 5.0 |
| 5.100.9 | **24411** | **0.0** | 115.8 | 90.4 | **0.0** | **0** | 12 | **0** | **0** | **0.0** | **0** | 4.2 |
| Average | **24197** | **0.0** | 91.0 | 81.0 | **0.0** | **0** | 9 | 10 | **0** | **0.0** | **0** | 4.3 |
| 5.100.10 | **42757** | **0.0** | 96.1 | 90.8 | **0.0** | **0** | 52 | 52 | **0** | **0.0** | **0** | 7.2 |
| 5.100.11 | **42545** | **0.0** | 103.2 | 110.8 | **0.0** | **0** | 51 | 74 | **0** | 22.8 | **0** | 9.8 |
| 5.100.12 | **41968** | **0.0** | 64.0 | 63.2 | **0.0** | **0** | 9 | 9 | **0** | 0.8 | **0** | 8.7 |
| 5.100.13 | **45090** | **0.0** | 68.2 | 80.0 | **0.0** | **0** | **0** | **0** | **0** | 15.2 | **0** | 7.7 |
| 5.100.14 | **42218** | **0.0** | 68.6 | 44.9 | **0.0** | **0** | **0** | **0** | **0** | **0.0** | **0** | 6.1 |
| 5.100.15 | **42927** | **0.0** | 27.8 | 36.7 | **0.0** | **0** | **0** | **0** | **0** | **0.0** | **0** | 6.1 |
| 5.100.16 | **42009** | **0.0** | 105.0 | 137.6 | **0.0** | **0** | **0** | **0** | **0** | **0.0** | **0** | 6.1 |
| 5.100.17 | **45020** | **0.0** | 109.3 | 71.2 | **0.0** | **0** | 10 | **0** | **0** | **0.0** | **0** | 7.5 |
| 5.100.18 | **43441** | **0.0** | 140.1 | 151.0 | **0.0** | **0** | **0** | 60 | **0** | 6.9 | **0** | 8.8 |
| 5.100.19 | **44554** | **0.0** | 60.2 | 81.6 | **0.0** | **0** | **0** | 25 | **0** | 13.5 | **0** | 7.8 |
| Average | **43253** | **0.0** | 84.3 | 86.8 | **0.0** | **0** | 12 | 22 | **0** | 5.9 | **0** | 7.6 |
| 5.100.20 | **59822** | **0.0** | 45.6 | 30.1 | **0.0** | **0** | **0** | **0** | **0** | **0.0** | **0** | 6.5 |
| 5.100.21 | **62081** | **0.0** | 144.1 | 120.7 | **0.0** | **0** | **0** | **0** | **0** | 9.4 | **0** | 6.9 |
| 5.100.22 | **59802** | **0.0** | 105.5 | 108.7 | **0.0** | **0** | **0** | 48 | **0** | 7.6 | **0** | 8.7 |
| 5.100.23 | **60479** | **0.0** | 75.0 | 90.2 | **0.0** | **0** | 1 | 1 | **0** | 0.2 | **0** | 10.1 |
| 5.100.24 | **61091** | **0.0** | 94.6 | 85.4 | **0.0** | **0** | 36 | 12 | **0** | **0.0** | **0** | 6.4 |
| 5.100.25 | **58959** | **0.0** | 54.8 | 72.7 | **0.0** | **0** | **0** | 22 | **0** | **0.0** | **0** | 9.1 |
| 5.100.26 | **61538** | **0.0** | 100.8 | 68.9 | **0.0** | **0** | **0** | **0** | **0** | **0.0** | **0** | 8.0 |
| 5.100.27 | **61520** | **0.0** | 109.0 | 91.9 | **0.0** | **0** | 31 | **0** | **0** | 5.8 | **0** | 9.2 |
| 5.100.28 | **59453** | **0.0** | 151.8 | 157.9 | **0.0** | **0** | **0** | **0** | **0** | **0.0** | **0** | 8.1 |
| 5.100.29 | **59965** | **0.0** | 36.8 | 21.4 | **0.0** | **0** | 5 | 5 | **0** | 0.8 | **0** | 8.5 |
| Average | **60471** | **0.0** | 91.8 | 84.8 | **0.0** | **0** | 7 | 9 | **0** | 2.4 | **0** | 8.2 |

Finally, large Glover and Kochenberger (GK) instances were solved and compared to eight heuristic algorithms from the literature in terms of average error percent (2) gap against best known profit from the literature. Compared algorithms include ACOwD, IWOA, TPTEA, harmony search based algorithm NBHS2 proposed in [20], evolutionary algorithm with logic gates LGEA [16], shuffled complex evolution algorithm SCEcr [17], hyper-heuristic inspired CF-LAS [44] and BCSA – binary cuckoo search algorithm [21]. Table 4 is colour coded from red (worst average error %) to the best average error percent, in green, for each problem instance with dashes (-) representing scores that were not available. Compared to 8 other algorithms in the literature, ICAwICA shows competitive results, coming in second place for gk01-gk09 and in top three for gk10 and in fourth place for the largest gk11 instance. The best achieved error percentage along with the average time $t_{avg}(s)$ have been included for reference. The proposed algorithm performs well on medium to large MKP instances, however, struggles on very large instances (gk11). Further investigation needs to be conducted to improve performance on the most complex benchmarks.

**Table 3.** Algorithm comparison and computational results for small OR-library instances with n=100 and m = 10. With bold corresponding to the best absolute error. Average derived from 30 independent runs.

| Problem Instance | Optimum | Average absolute error | | | | Best absolute error | | | | ICAwICA (this work) | | |
|---|---|---|---|---|---|---|---|---|---|---|---|---|
| | | TPTEA [15] | SACRO-BPPSO-TVAC [19] | SACRO-CBPPSO-TVAC [19] | QPSO* [18] | TPTEA [15] | SACRO-BPPSO-TVAC [19] | SACRO-CBPPSO-TVAC [19] | QPSO* [18] | Avg | Best | $t_{avg}(s)$ |
| 10.100.0 | **23064** | **0.0** | 50.8 | 55.0 | **0.0** | **0** | **0** | **0** | **0** | 0.0 | 0 | 4.3 |
| 10.100.1 | **22801** | **0.0** | 150.0 | 168.0 | **0.0** | **0** | **0** | **0** | **0** | 0.0 | 0 | 4.3 |
| 10.100.2 | **22131** | **0.0** | 182.3 | 138.8 | **0.0** | **0** | **0** | **0** | **0** | 0.0 | 0 | 4.1 |
| 10.100.3 | **22772** | **0.0** | 152.4 | 163.3 | **0.0** | **0** | **0** | **0** | **0** | 0.0 | 0 | 4.7 |
| 10.100.4 | **22751** | **0.0** | 136.1 | 144.1 | **0.0** | **0** | **0** | **0** | **0** | 0.0 | 0 | 4.8 |
| 10.100.5 | **22777** | **0.0** | 154.1 | 165.0 | **0.0** | **0** | **0** | 38 | **0** | 4.3 | 0 | 4.9 |
| 10.100.6 | **21875** | **0.0** | 98.8 | 91.7 | **0.0** | **0** | **0** | **0** | **0** | 0.0 | 0 | 4.2 |
| 10.100.7 | **22635** | **0.0** | 176.8 | 187.2 | **0.0** | **0** | **0** | **0** | **0** | 0.0 | 0 | 4.2 |
| 10.100.8 | **22511** | **0.0** | 149.7 | 169.0 | **0.0** | **0** | **0** | **0** | **0** | 0.0 | 0 | 4.2 |
| 10.100.9 | **22702** | **0.0** | 234.6 | 223.0 | **0.0** | **0** | **0** | **0** | **0** | 0.0 | 0 | 3.8 |
| Average | **22602** | **0.0** | 148.6 | 150.5 | **0.0** | **0** | **0** | 4 | **0** | 0.4 | 0 | 4.3 |
| 10.100.10 | **41395** | **0.0** | 109.4 | 121.8 | **0.0** | **0** | **0** | **0** | **0** | 0.5 | 0 | 8.7 |
| 10.100.11 | **42344** | **0.0** | 122.7 | 119.1 | **0.0** | **0** | **0** | **0** | **0** | 1.9 | 0 | 8.3 |
| 10.100.12 | **42401** | **0.0** | 119.4 | 108.9 | **0.0** | **0** | **0** | **0** | **0** | 0.0 | 0 | 7.5 |
| 10.100.13 | **45624** | **0.0** | 202.0 | 198.7 | **0.0** | **0** | **0** | **0** | **0** | 1.7 | 0 | 8.2 |
| 10.100.14 | **41884** | **0.0** | 142.0 | 135.1 | **0.0** | **0** | **0** | **0** | **0** | 0.0 | 0 | 8.0 |
| 10.100.15 | **42995** | **0.0** | 118.4 | 158.7 | **0.0** | **0** | **0** | **0** | **0** | 0.0 | 0 | 7.1 |
| 10.100.16 | **43574** | **0.0** | 194.1 | 176.0 | 21.0 | **0** | 15 | **0** | 21 | 1.5 | 0 | 8.8 |
| 10.100.17 | **42970** | **0.0** | 102.9 | 94.7 | **0.0** | **0** | **0** | **0** | **0** | 0.0 | 0 | 6.8 |
| 10.100.18 | **42212** | **0.0** | 78.4 | 77.5 | **0.0** | **0** | **0** | **0** | **0** | 0.0 | 0 | 7.1 |
| 10.100.19 | **41207** | **0.0** | 150.8 | 165.2 | **0.0** | **0** | **0** | **0** | **0** | 1.4 | 0 | 7.7 |
| Average | **42661** | **0.0** | 134.0 | 135.6 | 2.1 | **0** | 2 | **0** | 2.1 | 0.7 | 0 | 7.8 |
| 10.100.20 | **57375** | **0.0** | 74.9 | 63.1 | **0.0** | **0** | **0** | **0** | **0** | 0.0 | 0 | 7.0 |
| 10.100.21 | **58978** | **0.0** | 139.1 | 106.7 | **0.0** | **0** | **0** | **0** | **0** | 5.6 | 0 | 9.6 |
| 10.100.22 | **58391** | **0.0** | 129.8 | 107.4 | **0.0** | **0** | **0** | **0** | **0** | 9.8 | 0 | 9.0 |
| 10.100.23 | **61966** | **0.0** | 112.8 | 89.4 | **0.0** | **0** | **0** | **0** | **0** | 0.0 | 0 | 8.2 |
| 10.100.24 | **60803** | **0.0** | 32.2 | 36.7 | **0.0** | **0** | **0** | **0** | **0** | 0.0 | 0 | 7.9 |
| 10.100.25 | **61437** | **0.0** | 160.7 | 150.9 | **0.0** | **0** | **0** | **0** | **0** | 25.3 | 0 | 8.9 |
| 10.100.26 | **56377** | **0.0** | 63.7 | 74.8 | **0.0** | **0** | **0** | **0** | **0** | 0.0 | 0 | 7.7 |
| 10.100.27 | **59391** | **0.0** | 173.2 | 150.3 | **0.0** | **0** | **0** | **0** | **0** | 0.0 | 0 | 8.1 |
| 10.100.28 | **60205** | **0.0** | 165.8 | 176.0 | **0.0** | **0** | **0** | **0** | **0** | 0.0 | 0 | 7.8 |
| 10.100.29 | **60633** | **0.0** | 76.9 | 83.4 | **0.0** | **0** | **0** | **0** | **0** | 0.0 | 0 | 7.1 |
| Average | **59556** | **0.0** | 112.9 | 103.9 | **0.0** | **0** | **0** | **0** | **0** | 4.1 | 0 | 8.1 |

**Table 4.** Algorithm comparison across large Glover and Kochenberger (GK) knapsack instances. Expressed as average error percentage gap % against best known profit. Colour coded from best gap (green) to worst gap (red) for any given dataset. With dash (-) representing results that are not available.

| Data set | Problem size (n x m) | Best known | ACOwD [11] | NBHS2 [20] | IWOA [22] | LGEA [16] | TPTEA [15] | SCEcr [17] | CF-LAS [44] | BCSA [21] | ICAwICA (this work) | | |
|---|---|---|---|---|---|---|---|---|---|---|---|---|---|
| | | | | | | | | | | | Average | Best | $t_{avg}(s)$ |
| gk01 | 100x15 | 3766 | 0.14% | 0.29% | 0.68% | 0.66% | 0.00% | 0.76% | 0.31% | 0.23% | 0.00% | 0.00% | 16.8 |
| gk02 | 100x25 | 3958 | 0.05% | 0.30% | - | 0.55% | 0.00% | 1.06% | 0.36% | 0.27% | 0.05% | 0.03% | 19.4 |
| gk03 | 150x25 | 5656 | 0.26% | 0.55% | 0.85% | 0.97% | 0.06% | 0.91% | 0.37% | 0.17% | 0.12% | 0.11% | 62.5 |
| gk04 | 150x50 | 5767 | 0.17% | 0.45% | 0.89% | 1.02% | 0.01% | 1.48% | 0.45% | 0.15% | 0.07% | 0.05% | 84.4 |
| gk05 | 200x25 | 7561 | 0.21% | 0.44% | 0.94% | 1.32% | 0.01% | 0.73% | 0.24% | 0.18% | 0.09% | 0.04% | 145.7 |
| gk06 | 200x50 | 7680 | 0.26% | 0.52% | 0.77% | 1.05% | 0.08% | 1.14% | 0.46% | 3.54% | 0.13% | 0.12% | 247.7 |
| gk07 | 500x25 | 19221 | 0.20% | 0.26% | 1.09% | 1.08% | 0.04% | 0.46% | 0.13% | 0.70% | 0.11% | 0.07% | 280.3 |
| gk08 | 500x50 | 18806 | 0.22% | 0.56% | 0.85% | - | 0.06% | 0.67% | 0.20% | 0.77% | 0.12% | 0.08% | 357.8 |
| gk09 | 1500x25 | 58091 | 0.18% | 0.27% | 1.54% | 1.08% | 0.02% | 1.78% | 1.77% | 0.98% | 0.14% | 0.09% | 1611.0 |
| gk10 | 1500x50 | 57295 | 0.20% | 0.54% | 0.80% | 1.01% | 0.04% | 0.36% | 0.10% | - | 0.18% | 0.12% | 2219.1 |
| gk11 | 2500x100 | 95238 | 0.32% | 0.64% | 1.07% | 1.13% | 0.07% | 0.30% | 0.09% | - | 0.31% | 0.24% | 7200.6 |

# 5. Conclusions and future work

This work solves multidimensional NP-hard 0-1 knapsack problems using an Imperialist Competitive Algorithm for the first time. Furthermore, a new Independence operator has been presented that allows each of the colonies a free will to choose between moving closer to their own imperialist or another empire's imperialist. Moreover, a constraint assimilation process has been proposed that eliminates the need for a revolution operation in classical ICA.

In order to test the proposed algorithm, an independence operator was investigated, it was concluded that search performance improves when the operator is used. Next, algorithm was compared to multiple state-of-the-art algorithms across 101 MKP instances. Algorithm was able to achieve an optimal solution in all small instances and showed very competitive results in large GK instances.

The proposed algorithm can be improved in multiple ways. First, instead of simply iterating over all possibilities, a more efficient selection process based on heuristics can be explored. Similarly, the proposed independence operator is slow as it is required to assimilate to all imperialists, smarter selection of top imperialists can be implemented. Furthermore, although this paper focuses on solving binary MKP instances, the proposed algorithm is not limited to binary optimization and therefore applications to other constrained discrete problems is something to be considered.